\documentclass[10pt,twocolumn,letterpaper]{article}

\usepackage{cvpr}
\usepackage{times}
\usepackage{epsfig}
\usepackage{graphicx}
\usepackage{amsmath}
\usepackage{amssymb}
\usepackage{subcaption}
\usepackage{xcolor}
\usepackage[titletoc,title]{appendix}

\usepackage{multirow}
\usepackage{makecell}

\usepackage[pagebackref=true,breaklinks=true,colorlinks,bookmarks=false]{hyperref}

\cvprfinalcopy 

\def\yolofish{RAPiD}
\def\newhabbof{CEPDOF}
\def\rotbb{RBB}
\def\hitachi{Tamura {\it et al.}}
\def\shengye{Li {\it et al.}}
\def\mworiginal{MW}
\def\mw{MW-R}
\def\rroi{RRoI}
\def\hroi{HRoI}

\def\bbsymbol{\boldsymbol{\hat{b}}} 
\def\bbgtsymbol{\boldsymbol{b}} 
\def\bbnormsymbol{\boldsymbol{\hat{t}}} 

\def\Tsymbol{\widehat{T}}

\def\sigmoid{\mathrm{Sig}}
\def\bxpred{\widehat{b}_x}
\def\bypred{\widehat{b}_y}
\def\bwpred{\widehat{b}_w}
\def\bhpred{\widehat{b}_h}
\def\bthetapred{\widehat{b}_{\theta}}
\def\bconfpred{\widehat{b}_{\text{conf}}}
\def\txpred{\widehat{t}_x}
\def\typred{\widehat{t}_y}
\def\twpred{\widehat{t}_w}
\def\thpred{\widehat{t}_h}
\def\tthetapred{\widehat{t}_{\theta}}
\def\tconfpred{\widehat{t}_{\text{conf}}}
\def\bxgt{b_x}
\def\bygt{b_y}
\def\bwgt{b_w}
\def\bhgt{b_h}
\def\bthetagt{b_{\theta}}

\def\txgt{t_x}
\def\tygt{t_y}
\def\twgt{t_w}
\def\thgt{t_h}

\newif\ifcomments
\commentstrue
\commentsfalse

\ifcomments
    \def\ozan#1{{\color{green} [Ozan: #1]}}
    \def\ozanrep#1#2{{\color{red} \sout{#1}}{\color{green}\ #2}}
    \def\zhihao#1{{\color{orange} [Zhihao: #1]}}
    \def\zhihaorep#1#2{{\color{red} \sout{#1}}{\color{orange}\ #2}}
    \def\jlk#1{{\color{cyan} [JLK: #1]}}
    
    \def\piedit#1{{\color{magenta} [PI: #1]}}
    
\else 
    \def\ozan#1{}
    \def\ozanrep#1#2{#2}
    \def\zhihao#1{}
    \def\zhihaorep#1#2{#2}
    \def\jlk#1{}
    
    \def\piedit#1{}
    
\fi

\def\cvprPaperID{14} 

\ifcvprfinal\fi
\begin{document}

\title{RAPiD: Rotation-Aware People Detection in Overhead Fisheye Images}

\author{
Zhihao Duan, \ M. Ozan Tezcan, \ Hayato Nakamura, \ Prakash Ishwar, \ Janusz Konrad  
\thanks{This work was supported in part by ARPA-E under agreement DE-AR0000944 and by the donation of Titan GPUs from NVIDIA Corp.}\\
Boston University\\
{\tt\small \{duanzh, mtezcan, nhayato, pi, jkonrad\}@bu.edu}
}

\maketitle

\begin{abstract}
Recent methods for people detection in overhead, fisheye images either use radially-aligned bounding boxes to represent people, assuming people always {appear} along {image} radius or
{require significant} pre-/post-processing which 
{radically increases computational complexity}.
In this work, we 
{develop} an end-to-end rotation-aware people detection method, named \yolofish{}, that detects people using arbitrarily-oriented bounding boxes. Our fully-convolutional neural network directly regresses the angle of each bounding box using a periodic loss function, which 
{accounts} for 
{angle periodicities}. 
{We have also created} a new dataset\footnote{\href{http://vip.bu.edu/cepdof}{\tt vip.bu.edu/cepdof}} 
with spatio-temporal annotations of rotated bounding boxes, for people detection as well as other vision tasks in overhead fisheye videos. We show that our simple, yet effective method outperforms state-of-the-art results on three fisheye-image datasets. The source code for \yolofish{} is publicly available\footnote{\href{http://vip.bu.edu/rapid}{\tt vip.bu.edu/rapid}}.

\end{abstract}

\vspace{-1 ex}
\section{Introduction}


Occupancy sensing is an enabling technology for smart buildings of the future; knowing where and how many people are in a building is key for saving energy, space management and security (e.g., fire, active shooter). Various approaches to counting people have been developed to date, from virtual door tripwires to WiFi signal monitoring. Among those, video cameras combined with computer vision algorithms have proven most successful \cite{enzweiler2008survey,nguyen2016survey,brunetti2018survey_deep}. Typically, a wide-angle, standard-lens camera is side-mounted above the scene; multiple such cameras are used for large spaces. An alternative is to use a single overhead, fisheye camera with a 360$^\circ$ field of view (FOV). However, people detection algorithms developed for side-view, standard-lens images do not perform well on overhead, fisheye images due to their unique radial geometry and barrel distortions.

\begin{figure}[t]
    \begin{subfigure}[t]{0.15\textwidth}
        \centering
        \includegraphics[height=1in]{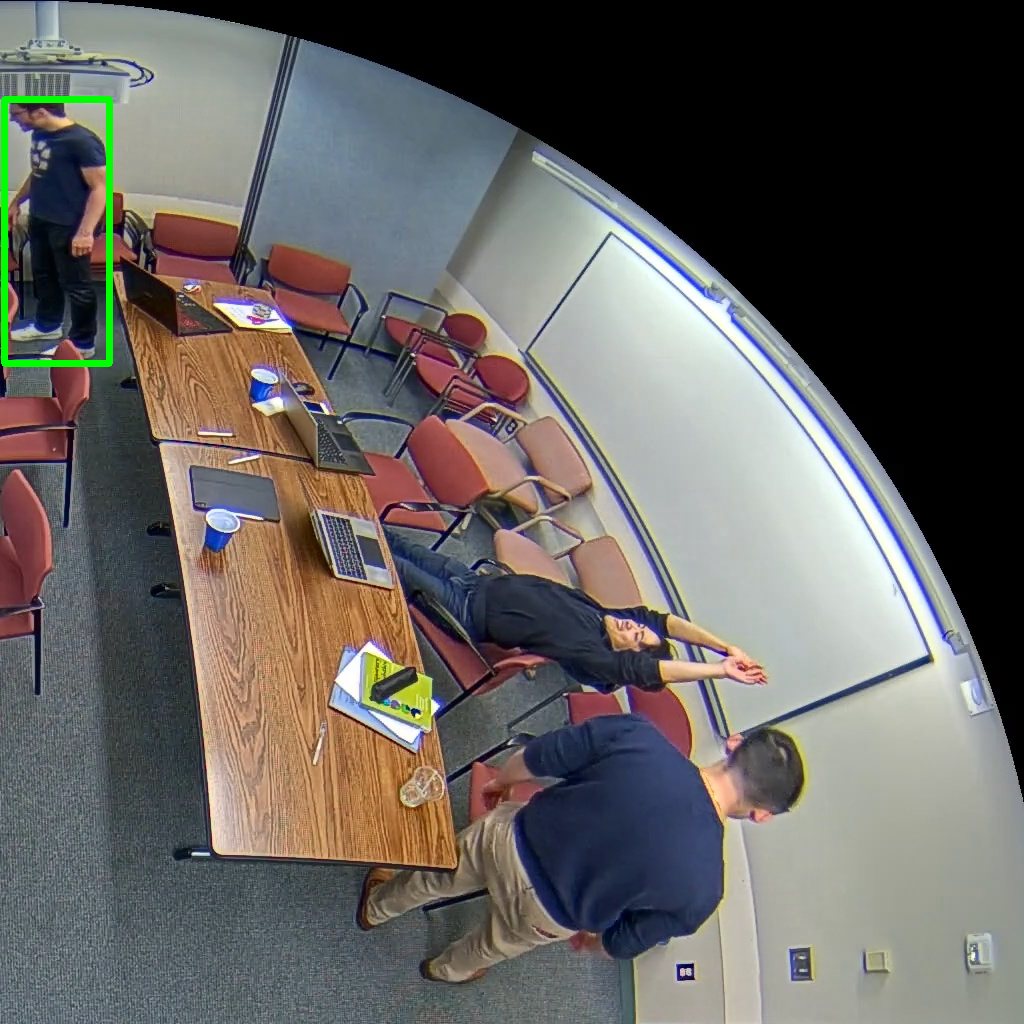}
        \caption{Axis-aligned}
        \label{fig:axisBB}
    \end{subfigure}
    \begin{subfigure}[t]{0.15\textwidth}
        \centering
        \includegraphics[height=1in]{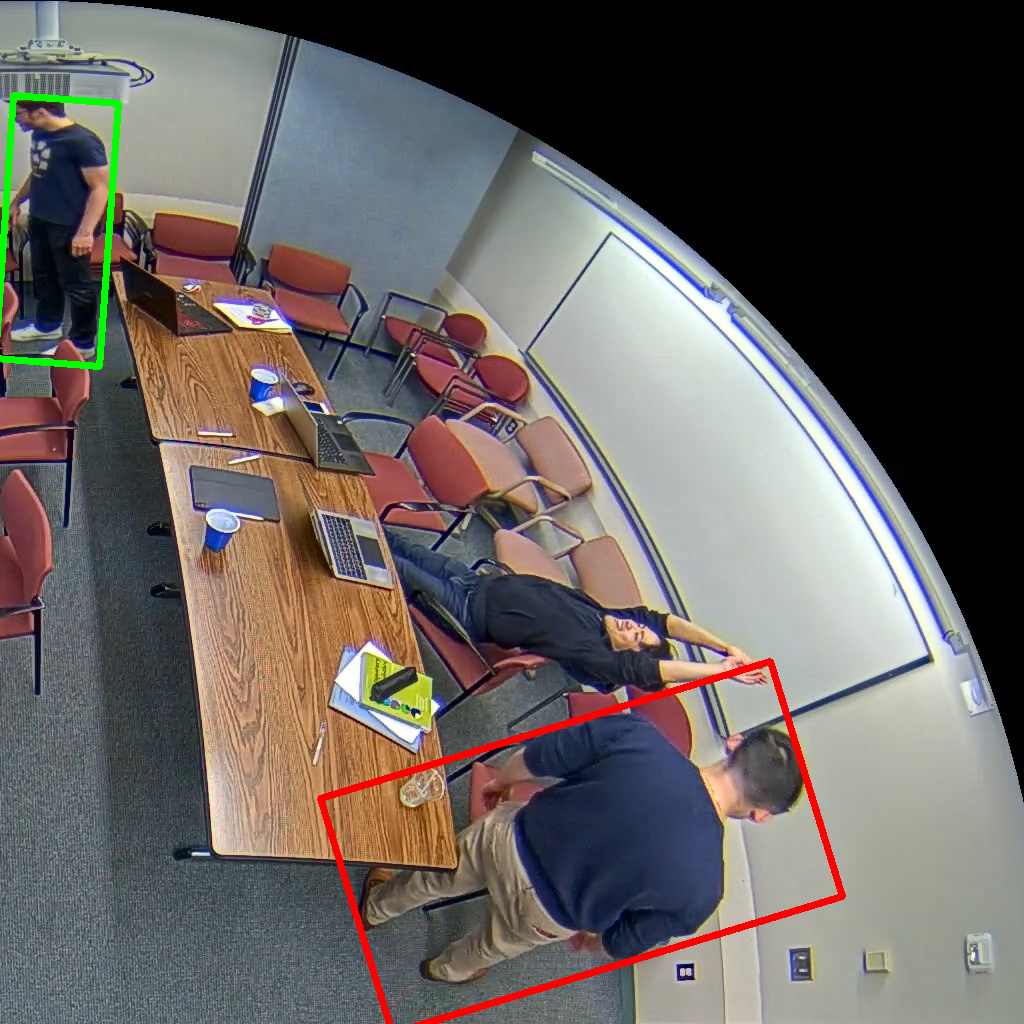}
        \caption{Radius-aligned}
        \label{fig:radBB}
    \end{subfigure}
    \begin{subfigure}[t]{0.15\textwidth}
        \centering
        \includegraphics[height=1in]{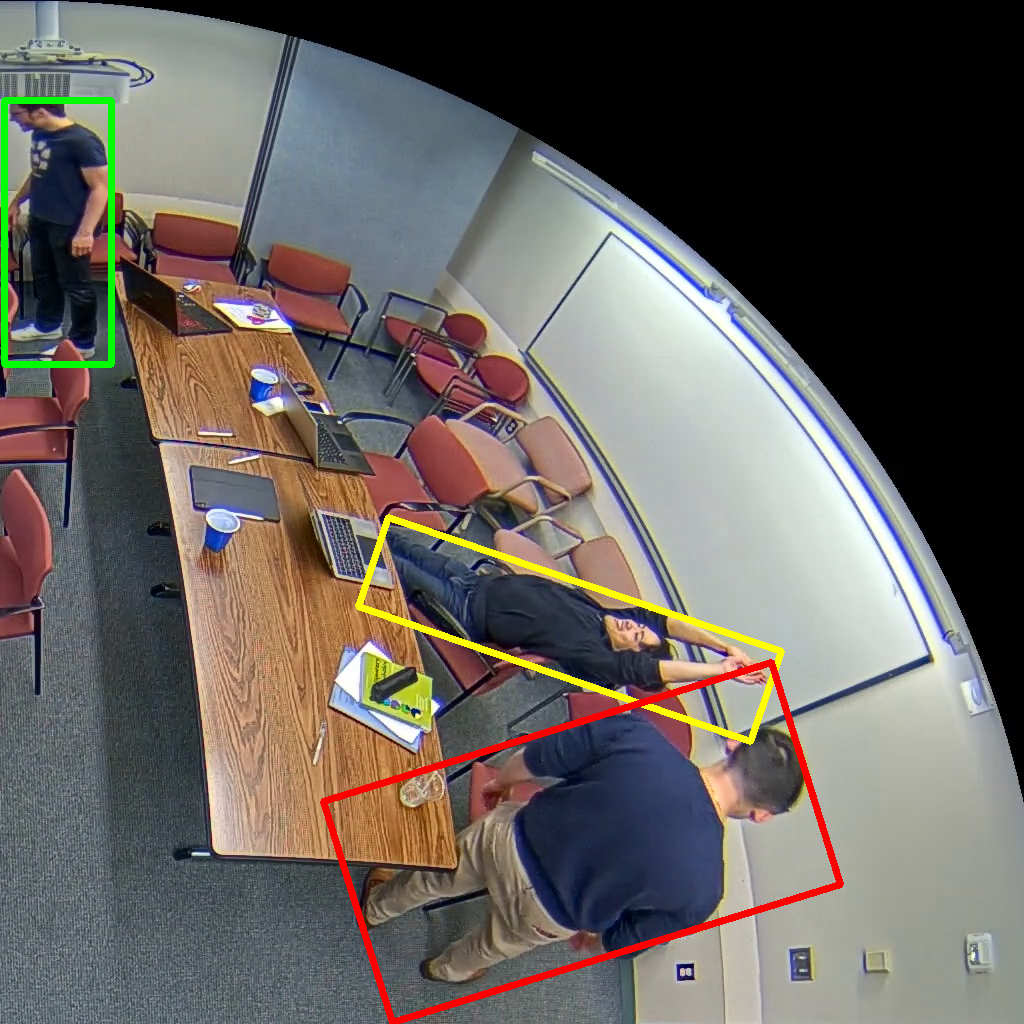}
        \caption{Human-aligned}
        \label{fig:rotBB}
    \end{subfigure}
    \vskip -0.2cm
    \caption{{Illustration of {\it typical} people-detection results on overhead, fisheye images (one quarter shown) for algorithms using various bounding-box orientation constraints; the human-aligned bounding boxes fit bodies most accurately. These are not outputs from any algorithms. See the text for discussion.}}
    \label{fig:BB}
    \vskip -0.7cm
\end{figure}

{In standard images, standing people usually appear in an upright position and algorithms that detect bounding boxes aligned with image axes, such as YOLO \cite{yolo}, SSD \cite{ssd} and R-CNN \cite{fasterrcnn}, work well. However, these algorithms perform poorly on overhead, fisheye images \cite{shengye}, usually missing non-upright bodies (Fig.~\ref{fig:axisBB}). In such images, standing people appear along image radius, due to the overhead placement of the camera, and rotated bounding boxes are needed. To accommodate this rotation, several people-detection algorithms, mostly YOLO-based, have been recently proposed \cite{Chiang2014HumanDI, Wang2017TemplateBP, Krams2017PeopleDI, shengye, hitachi, zhou2017orientationawarecnn}, each dealing differently with the radial geometry. For example, in one of the top-performing algorithms \cite{shengye}, the image is rotated in 15$^\circ$ steps and YOLO is applied to the top--center part of the image (where people usually appear upright) followed by post-processing. However, this requires 24-fold application of YOLO. Another recent algorithm \cite{hitachi} requires that bounding boxes be aligned with image radius, but often fails to detect non-standing poses (Fig.~\ref{fig:radBB}).}

In this paper, we introduce Rotation-Aware People Detection (\yolofish{}), a novel end-to-end people-detection algorithm for overhead, fisheye images.
\yolofish{} is a single-stage convolutional neural network that predicts {arbitrarily-rotated} bounding boxes (Fig.~\ref{fig:rotBB}) of people in a fisheye image. It extends the model proposed in YOLO \cite{yolo, yolov2, yolov3}, one of the most successful object detection algorithms for standard images.
In addition to 
{predicting} the center and size of a bounding box, \yolofish{} also predicts its angle.
{This is accomplished by} a periodic loss function 
{based on an extension of a common regression loss}. This allows us to predict the exact rotation of each bounding box in an image without any assumptions and additional computational complexity.
Since \yolofish{} is an end-to-end algorithm, we can train or fine-tune its weights on annotated fisheye images. Indeed, we show that such fine-tuning of a model trained on standard images significantly increases the performance.
{An additional aspect of this work, motivated by its focus on people detection, is the replacement of the common regression-based loss function used in multi-class object detection algorithms \cite{yolo, ssd, fastrcnn, fasterrcnn} with single-class object detection.}
The inference speed of \yolofish{} is nearly identical to that of YOLO since it is applied to each image only once without the need for pre-/post-processing.

We {evaluate} the performance of \yolofish{} on two publicly-available, people-detection datasets captured by overhead fisheye cameras, Mirror Worlds (\mworiginal{})\footnote{\href{http://www2.icat.vt.edu/mirrorworlds/challenge/index.html}{\tt www2.icat.vt.edu/mirrorworlds/challenge/index.html}} and HABBOF \cite{shengye}. Although these datasets cover a range of scenarios, they lack challenging cases such as unusual body poses, wearing a hoodie or hat, holding an object, carrying a backpack, strong occlusions, or low light. Therefore, we introduce a new dataset {named} Challenging Events for Person Detection from Overhead Fisheye images (\newhabbof{}) that includes such scenarios. In our evaluations, \yolofish{} outperforms state-of-the-art algorithms on all three datasets.

{The main contributions of this work can be summarized as follows:}

\begin{itemize}
  \item We propose an end-to-end neural network, which extends YOLO v3, for rotation-aware people detection in overhead fisheye images and demonstrate that our simple, yet effective approach, outperforms the state-of-the-art methods.
  
  
  \item We propose a continuous, periodic loss function for bounding-box angle that, unlike in previous methods, facilitates arbitrarily-oriented bounding boxes capable of handling a wide range of human-body poses.
  
  \item We introduce a new dataset for people detection from overhead, fisheye cameras that includes a range of challenges; it can be also useful for other tasks, such as people tracking and re-identification. 
  
\end{itemize}

\section{Related work} \label{sec:related}

\noindent\textbf{People detection using side-view standard-lens cameras: } Among traditional people-detection algorithms for standard cameras, the most popular ones are based on the histogram of oriented gradients (HOG) \cite{dalal2005hog} and aggregate channel features (ACF) \cite{dollar2014acf}. Recently, deep learning algorithms have demonstrated outstanding performance in object and people detection \cite{yolo, ssd, dssd, fastrcnn, fasterrcnn, maskrcnn}. These algorithms can be divided into two categories: two-stage methods and one-stage methods. Two-stage methods, such as R-CNN and its variants \cite{fastrcnn, fasterrcnn, maskrcnn}, consist of a Region Proposal Network (RPN) {which predicts the Region of Interest (ROI)} and a network head 
{refines the bounding boxes}. One-stage methods, such as {variants of} SSD
\cite{ssd, dssd} and YOLO
\cite{yolo, yolov2, yolov3}, could be viewed as independent RPNs. Given an input image, one-stage methods directly regress bounding boxes 
through CNNs. Recently, attention has focused on fast one-stage detectors \cite{m2det, efficientdet} and anchor-free detectors \cite{fcos, atss}.


\noindent\textbf{Object detection using rotated bounding boxes:}
Detection of rotated bounding boxes has been widely studied in text detection and aerial image analysis \cite{ma2018rrpn,roitransformer,r3det,qian2019rsdet}. RRPN \cite{ma2018rrpn} is a two-stage object detection algorithm which uses rotated anchor boxes and a rotated region-of-interest (\rroi{}) layer. RoI-Transformer \cite{roitransformer} extended this idea by first computing a horizontal region of interest (\hroi{}) and then learning the warping from \hroi{} to \rroi{}. R$^3$Det \cite{r3det} proposed a single-stage rotated bounding box detector by using a feature refinement layer to solve feature misalignment occurring between the region of interest and the feature, a common problem of single-stage methods.
{In an alternative approach, Nosaka {\it et al.} \cite{nosaka2018orientation} used orientation-aware convolutional layers \cite{zhou2017orientationawarecnn} to handle the bounding box orientation and a smooth $L1$ loss for angle regression.}
All of these methods use a {5-component vector} for rotated bounding boxes (coordinates of the center, width, height and rotation angle) with the angle defined in $[\frac{-\pi}{2}, 0]$ range and a traditional regression loss. Due to symmetry, a rectangular bounding box having width $b_w$, height $b_h$ and angle $\theta$ is indistinguishable from one having width $b_h$, height $b_w$ and angle $(\theta-\pi/2)$. Hence a standard regression loss, which does not account for this, may incur a large cost even when the prediction is close to the ground truth, e.g., if the ground-truth annotation is $(b_x, b_y, b_h, b_w, -4\pi/10)$, a prediction $(b_x, b_y, b_w, b_h, 0)$ may seem far from the ground truth, but is not so since the ground truth is equivalent to $(b_x, b_y, b_w, b_h, \pi/10)$.
%
%
%
%
RSDet \cite{qian2019rsdet} addresses this by introducing a modulated rotation loss.

\begin{figure*}[t]
\begin{center}
\includegraphics[width=1\linewidth]{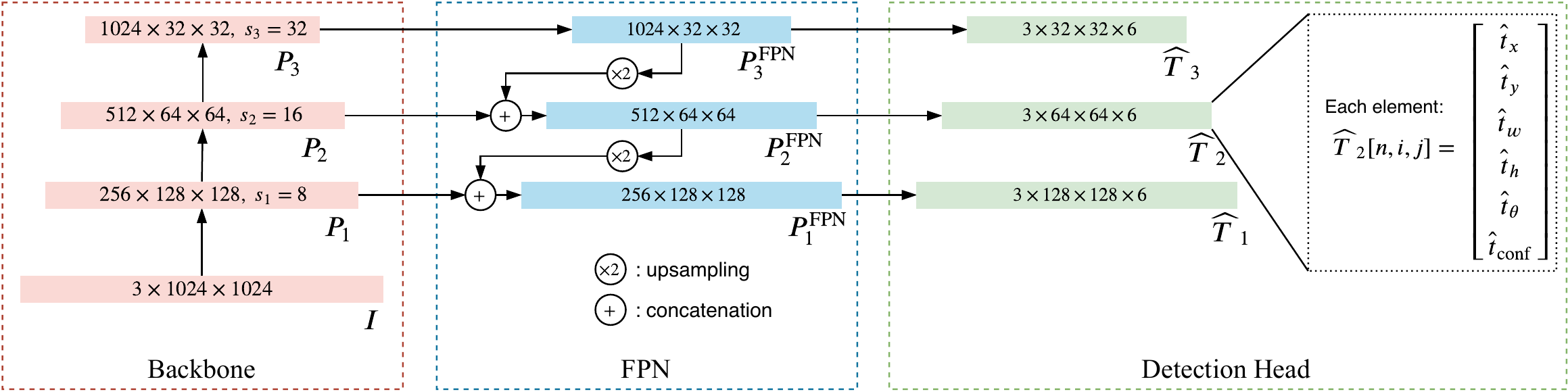}
\end{center}
\caption{\yolofish{} architecture. Following the paradigm of one-stage detectors, our model contains a backbone, FPN, and detection head (bounding-box regression network). In the diagram, each arrow represents multiple convolutional layers and the colored rectangles represent multi-dimensional matrices, i.e., feature maps, 
{whose dimensions correspond to input image of size $h \times w = 1,024 \times 1,024$.}
}
\label{fig:yolofisheye}
\vglue -0.3cm
\end{figure*}

\noindent\textbf{People detection in overhead, fisheye images:} 
People detection using overhead, fisheye cameras is an emerging area with sparse literature. In some approaches, traditional people-detection algorithms such as HOG and LBP have been applied to fisheye images with slight modifications to account for fisheye geometry \cite{Wang2017TemplateBP, Chiang2014HumanDI, saito2011people, Krams2017PeopleDI}. For example, Chiang and Wang \cite{Chiang2014HumanDI} rotated each fisheye image in small angular steps and extracted HOG features from the top-center part of the image. Subsequently, they applied SVM classifier to detect people. In another algorithm, Krams and Kiryati \cite{Krams2017PeopleDI} trained an ACF classifer on side-view images and dewarped the ACF features extracted from the fisheye image for person detection.

Recently, CNN-based algorithms have been applied to this problem as well. Tamura {\it et al.} introduced a rotation-invariant version of YOLO \cite{yolo} by training the network on a rotated version of the COCO dataset \cite{coco}. The inference stage in their method relies on the assumption that bounding boxes in a fisheye image are aligned with the image radius. 
Another YOLO-based algorithm \cite{seidel2018improved} applies YOLO to dewarped versions of overlapping windows extracted from a fisheye image.
Li {\it et al.} \cite{shengye} rotate each fisheye image in 15$^\circ$ steps and apply YOLO only to the upper-center part of the image where people usually appear upright. Subsequently, they apply post-processing to remove multiple detections of the same person. Although their algorithm is very accurate, it is computationally complex as it applies YOLO 24 times to each image.


{In this work, we introduce an angle-aware loss function to predict the exact angle of bounding boxes without any additional assumptions. We also change the commonly-used representation of rotated bounding boxes to overcome the symmetry problem (Section~\ref{sec:angle_range}).}

\section{Rotation-Aware People Detection (\yolofish{})}

We propose \yolofish{}, a new CNN that, in addition to the location and size, also estimates the angle of each bounding box in an overhead, fisheye image. During training, \yolofish{} includes a rotation-aware regression loss to account for these angles.
%
%
\yolofish{}'s design has been largely motivated by YOLO. Below,
we explain this design in detail and we highlight the concepts we borrowed from YOLO as well as novel ideas that we proposed.

\textbf{Notation:} 
We use $\bbgtsymbol{} = (\bxgt{}, \bygt{}, \bwgt{}, \bhgt{}, \bthetagt{}) \in \mathbb{R}^5$ to denote a ground-truth bounding box, where $\bxgt{}, \bygt{}$ are the coordinates of the bounding box center; $\bwgt{}, \bhgt{}$ are the width and height and $\bthetagt{}$ is the angle by which the bounding box is rotated clockwise. Similarly $\bbsymbol{} = (\bxpred{}, \bypred{}, \bwpred{}, \bhpred{}, \bthetapred{}, \bconfpred{}) \in \mathbb{R}^6$ denotes a predicted bounding box, where the additional element $\bconfpred{}$ denotes the confidence score of the prediction. All the angles used in the paper are in radians.


\subsection{Network Architecture}

Our object-detection network can be divided into three stages: backbone network, feature pyramid network (FPN) \cite{fpn}, and bounding box regression network, also known as the detection head:
\begin{equation}
  \begin{aligned}
    P_1, P_2, P_3 &= \text{Backbone}(I) \\
    P_1^{\text{fpn}}, P_2^{\text{fpn}}, P_3^{\text{fpn}} &= \text{FPN}(P_1, P_2, P_3)\\
    \Tsymbol_k &= \text{Head}_k(P_k^{\text{fpn}}) \quad \forall k = 1,2,3 \\ 
  \end{aligned}
\end{equation}
where $I \in [0,1]^{3\times h\times w}$ is the input image, $\{P_k\}_{k=1}^3$ denotes a multi-dimensional feature matrix and $\{\Tsymbol_k\}_{k=1}^3$ denotes a list of predicted bounding boxes in transformed notation (the relationship between $\Tsymbol$ and $\bbsymbol{}$ will be defined soon -- see equation (\ref{eq:inference})) at three levels of resolution. Fig.~\ref{fig:yolofisheye} shows the overall \yolofish{} architecture, while below we describe each stage in some depth. For more details, interested readers are referred to \cite{yolov3}.

\noindent\textbf{Backbone:} The backbone network, also known as the feature extractor, takes an input image $I$ and outputs a list of features $(P_1, P_2, P_3)$ from different parts of the network. The main goal is to extract features at different spatial resolutions ($P_1$ being the highest and $P_3$ being the lowest). By using this multi-resolution pyramid, we expect to leverage both the low-level and high-level information extracted from the image.



\noindent\textbf{Feature Pyramid Network (FPN):} The multi-resolution features computed by the backbone are fed into $FPN$ in order to extract features related to object detection, denoted $(P_1^{\text{fpn}}, P_2^{\text{fpn}}, P_3^{\text{fpn}})$. We expect $P_1^{\text{fpn}}$ to contain information about small objects and $P_3^{\text{fpn}}$ -- about large objects. 


\noindent\textbf{Detection Head:} After FPN, a separate CNN is applied to each feature vector $P_k^{\text{FPN}}, k \in \{1,2,3\}$ to produce a transformed version of bounding-box predictions, denoted $\Tsymbol_k$ -- a 4-dimensional matrix 
with $\langle 3, h/s_k, w/s_k, 6 \rangle$ dimensions. The first dimension indicates that there are three anchor boxes being used in $\Tsymbol_k$, the second and third dimensions denote the prediction grid, {where $h \times w$ is the resolution of the input image and $s_k$ is the stride at resolution level $k$ as shown in Fig.~\ref{fig:yolofisheye}, } 
and the last dimension denotes a transformed version of the predicted bounding box for each grid cell.
{We denote the $n^{th}$ transformed bounding box prediction of $Head_k$ in grid cell
$(i, j)$ as $\Tsymbol{}{}_k [n,i,j] = (\txpred{}, \typred{}, \twpred{}, \thpred{}, \tthetapred{}, \tconfpred{})$ from which a bounding-box prediction can be computed as follows:}
%
\begin{equation}
  \label{eq:inference}
  \begin{aligned}
    \bxpred{} &= s_{k}
    \left ( j + \sigmoid(\txpred{}) \right ), & \bwpred &= w_{k,n}^{\text{anchor}} e^{\twpred{}}\\ 
    \bypred{} &= s_{k} \left ( i + \sigmoid(\typred{}) \right ), & \bhpred{} &= h_{k,n}^{\text{anchor}} e^{\thpred{}}\\ 
    \bthetapred{} &= \alpha \ \sigmoid(\tthetapred{}) - \beta, & \bconfpred{} &= \sigmoid(\tconfpred{})
  \end{aligned}
\end{equation}
where $\sigmoid(\cdot)$ is the logistic (sigmoid) activation function and $w_{k,n}^{\text{anchor}}$ and $h_{k,n}^{\text{anchor}}$ are the width and height of the $n^{th}$ anchor box for $Head_k$.
Note, that angle prediction $\bthetapred{}$ is limited to range $[-\beta, \alpha-\beta]$ (\ref{eq:inference}). In Section~\ref{sec:angle_range} below, we discuss the selection of $\alpha$ and $\beta$ values.

\subsection{Angle-Aware Loss Function}\label{sec:loss}

Our loss function is inspired by that used in YOLOv3 \cite{yolov3}, with an additional bounding-box rotation-angle loss:
\begin{equation} \label{eq:loss}
  \begin{aligned}
    \mathfrak{L} &= \sum_{\bbnormsymbol{} \in \Tsymbol{}^{\text{pos}}} \text{BCE}(\sigmoid(\txpred{}), \txgt{}) + \text{BCE}(\sigmoid(\typred{}), \tygt{}) \\
    &+ \sum_{\bbnormsymbol{} \in \Tsymbol^{\text{pos}}} (\sigmoid(\twpred{}) - \twgt{})^2 + (\sigmoid(\thpred{}) - \thgt{})^2 \\
    &+ \sum_{\bbnormsymbol{} \in \Tsymbol^{\text{pos}}} \ell_{\text{angle}}(\bthetapred{}, \bthetagt{}) \\
    &+ \sum_{\bbnormsymbol{} \in \Tsymbol^{\text{pos}}} \text{BCE}(\sigmoid(\tconfpred{}),1) + \sum_{\bbnormsymbol{} \in \Tsymbol^{\text{neg}}} \text{BCE}(\sigmoid(\tconfpred{}),0)
  \end{aligned}
\end{equation}
where $BCE$ denotes binary cross-entropy,  $\ell_{\text{angle}}$ is a new angle loss function that we propose in the next section, $\Tsymbol^{\text{pos}}$ and $\Tsymbol^{\text{neg}}$ are positive and negative samples from the predictions, respectively, as described in YOLOv3, $\bthetapred{}$ is calculated in equation (\ref{eq:inference}) and $\txgt{}, \tygt{}, \twgt{}, \thgt{}$ are calculated from the ground truth as follows:
\begin{equation}
  \begin{aligned}
    \txgt{} &= \frac{\bxgt{}}{s_{k}} - \Bigg\lfloor \frac{\bxgt{}}{s_{k}} \Bigg\rfloor, & \twgt &= \ln\left ( \frac{\bwgt{}}{w_{k,n}^{\text{anchor}}} \right )\\ 
    \tygt{} &= \frac{\bygt{}}{s_{k}} - \Bigg\lfloor \frac{\bygt{}}{s_{k}} \Bigg\rfloor, & \thgt{} &= \ln\left ( \frac{\bhgt{}}{h_{k,n}^{\text{anchor}}} \right )
  \end{aligned}
\end{equation}
Note, that we do not use the category-classification loss since we use only one class (person) in our problem.

{Traditionally, regression functions based on $L1$ or $L2$ distance are used for angle prediction \cite{ma2018rrpn,roitransformer,r3det}. However, these metrics do not consider the periodicity of the angle and might result in misleading cost values due to symmetry in the parameterization of rotated bounding boxes.
We solve these issues by using a periodic loss function and changing the parameterization, respectively.}

\subsubsection{Periodic Loss for Angle Prediction} \label{sec:angle_loss}

\begin{figure}[t]
  \begin{center}
  \includegraphics[width=1\linewidth]{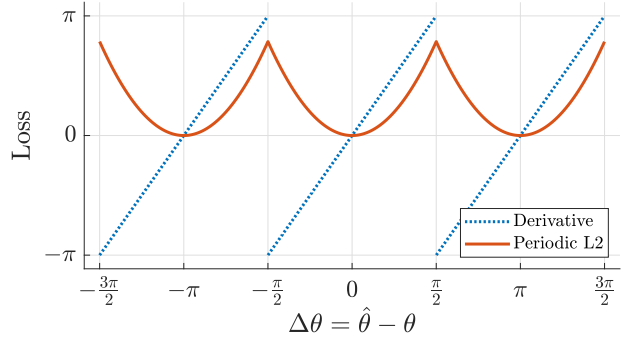}
  \end{center}
  \vspace{-2 ex}
  \caption{Periodic loss function with $L2$ norm as regressor and its derivative}
  \label{fig:pL2}
\end{figure}

%
%
Since a bounding box remains identical after rotation by $\pi$, the angle loss function must satisfy $\ell_{\text{angle}}(\widehat{\theta}, \theta) = \ell_{\text{angle}}(\widehat{\theta} + \pi, \theta)$, i.e., must be a $\pi$-periodic function with respect to $\widehat{\theta}$.

We propose a new, periodic angle loss function:
\begin{equation}
  \ell_{\text{angle}}(\hat{\theta},\theta) = f( \text{mod}(\hat{\theta}-\theta - \frac{\pi}{2}, \pi) - \frac{\pi}{2})
\end{equation}
where $\text{mod}(\cdot)$ denotes the modulo operation and $f$ is any symmetric regression function such as $L1$ or $L2$ norm. Since $\frac{\partial}{\partial x} \text{mod}(x,\cdot) = 1$, the derivative of this loss function with respect to $\hat{\theta}$ can be calculated as follows,
%
\begin{equation}\label{eq:angleloss_derivative}
  \ell_{\text{angle}}^\prime(\hat{\theta},\theta) = 
  f^\prime(\text{mod}(\hat{\theta}-\theta - \frac{\pi}{2}, \pi) - \frac{\pi}{2})
\end{equation}
{except for angles such that $\hat{\theta}-\theta=(k\pi + \pi/2)$ for integer $k$, where $\ell_{\text{angle}}$ is non-differentiable}. However, we can ignore these angles during backpropagation as is commonly done for other non-smooth functions, such as $L1$ distance.
Fig.~\ref{fig:pL2} shows an example plot of $\ell_{\text{angle}}(\hat{\theta}, \theta)$ with $L2$ distance as well as its derivative with respect to $\Delta\theta=\hat{\theta}-\theta$. 

\subsubsection{Parameterization of Rotated Bounding Boxes}
\label{sec:angle_range}

In most of the previous work on rotated bounding-box (\rotbb{}) detection, $[-\frac{\pi}{2},0]$ range is used for angle representation.
{This ensures that all \rotbb{}s can be uniquely expressed as $(b_x, b_y, b_w, b_h, b_\theta)$ where $b_\theta \in [-\frac{\pi}{2},0]$. However, as discussed in Section~\ref{sec:related} and also in \cite{qian2019rsdet}, this approach might lead to a large cost even when the prediction is close to the ground truth due to the symmetry of the representation, i.e., $(b_x, b_y, b_w, b_h, b_\theta) = (b_x, b_y, b_h, b_w, b_\theta-\pi/2)$.}
We address this by enforcing the following rule in our ground-truth annotations: $b_w < b_h$ and extending the ground-truth angle range to $[-\frac{\pi}{2},\frac{\pi}{2})$ to be able represent all possible \rotbb{}s.
For bounding boxes that are exact squares, a rare situation, we simply decrease a random side by 1 pixel.
Under this rule, each bounding box will correspond to a unique 5-D vector representation.

\begin{figure}[t]
  \begin{center}
    \includegraphics[width=0.6\linewidth]{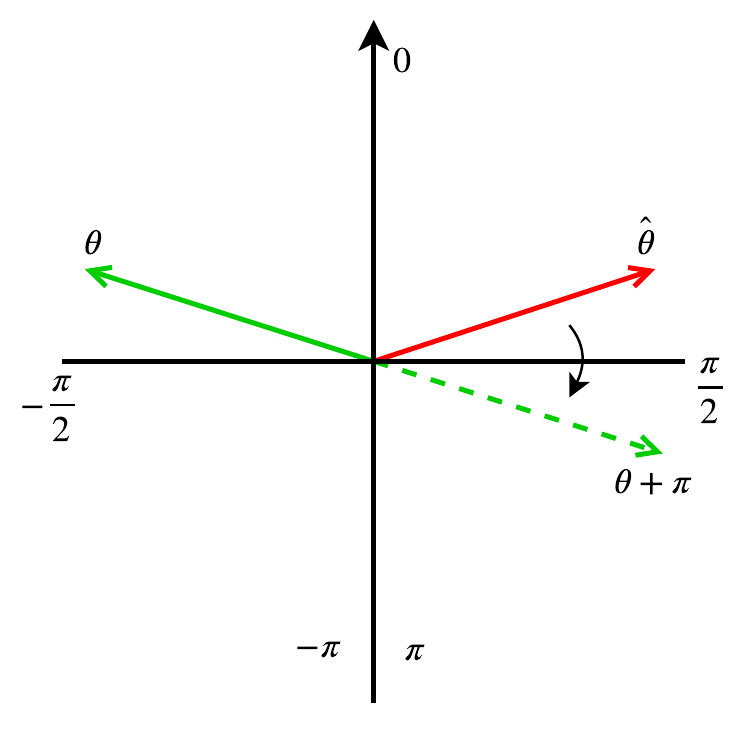}
  \end{center}
  \vspace{-3 ex}
  \caption{Illustration of the necessity to expand the predicted-angle value range. {Gradient descent applied to the predicted angle $\hat{\theta}$ (red arrow) may rotate it clockwise and away from the ground truth angle $\theta$ (green arrow). Since a bounding box at angle $\theta+\pi$ is the same as the one at $\theta$, we need to extend the angle range to include $\theta + \pi$ (dashed green arrow) otherwise $\widehat{\theta}$, pushed by the gradient, will stop at $\pi/2$.}}
  \label{fig:range}
\end{figure}

Given the fact that the ground-truth angle $\theta$ is defined in $[-\frac{\pi}{2},\frac{\pi}{2})$ range, it seems logical to force the predicted angle $\hat{\theta}$ to be in the same range by assigning $(\alpha, \beta) = (\pi, \pi/2)$ in equation (\ref{eq:inference}). However, this creates a problem for gradient descent when $\pi/2 < \hat{\theta} - \theta < \pi$ since the derivative of angle loss (\ref{eq:angleloss_derivative}) will be negative (Fig.~\ref{fig:pL2}). In this case, gradient descent will tend to increase $\hat{\theta}$ which will move it further away from the actual angle $\theta$. Clearly, the network should learn to estimate the angle as $\theta + \pi$ instead of $\theta$ (Fig.~\ref{fig:range}). To allow this kind of behavior, we extend the range of allowed angle predictions to $[-\pi,\pi)$ by assigning $(\alpha, \beta) = (2\pi, \pi)$. 

{Note that our new \rotbb{} parameterization will not have the symmetry problem explained above if the network eventually learns to predict the parametrization rule, $\bwpred{} \leq \bhpred{}$, which is very likely considering the fact that all ground-truth \rotbb{}s satisfy $\bwgt{} \leq \bhgt{}$.} Indeed, based on our experiments in Section~\ref{sec:angleloss_hist} we show that nearly all \rotbb{}s predicted by \yolofish{} satisfy $\bwpred{} \leq \bhpred{}$.

{In summary, by 1) defining $[-\frac{\pi}{2},\frac{\pi}{2})$ as the ground truth angle range and forcing ground truth $b_w < b_h$, 2) using our proposed periodic angle loss function, and 3) setting predicted angle range to be $(-\pi,\pi)$, our network can learn to predict arbitrarily-oriented \rotbb{}s without problems experienced by previous \rotbb{} methods. Based on the experimental results in Section~\ref{sec:anglelosstable}, we choose periodic L1 to be our angle loss function $\ell_{\text{angle}}$.}



\subsection{Inference}

During inference, an image $I \in \mathbb{R}^{3\times h\times w}$ is fed into the network, and three groups of bounding boxes (from three feature resolutions) are obtained.
A confidence threshold is applied to select the best bounding box predictions. 
After that, non-maximum suppression (NMS) is applied to remove redundant detections of the same person. 


\section{Experimental Results}





\begin{table*}[ht] \centering
\caption{Statistics of our new \newhabbof{} dataset in comparison with existing overhead fisheye image datasets. 
Since all fisheye images have a field of view with 1:1 aspect ratio, we only list one dimension (i.e., ``$1,056$ to $1,488$'' means frame resolution for different videos is between $1,056 \times 1,056$ and $1,488 \times 1,488$).
Note that the \mw{} dataset in this table is a subset of the original \mworiginal{} dataset that we annotated with bounding-box rotation angles.}
\vspace{-1 ex}
\begin{tabular}{|l|cccccc|}
\hline
Dataset  & \# of videos & Avg. \# of people & Max \# of people & \# of frames & Resolution     & FPS  \\ \hline
\mw{} & 19           & 2.6               & 6                & 8,752        & 1,056 to 1,488 & 15   \\
HABBOF & 4            & 3.5               & 5                & 5,837        & 2,048          & 30   \\
\newhabbof{} & {8}            & {6.8}               & 13               & {25,504}       & 1,080 to 2,048 & 1-10 \\ \hline
\end{tabular}
\label{table:stats}
\end{table*}

\begin{table*}[ht] \centering
\caption{\zhihaorep{}{Scenario details and statistics of individual videos in {\newhabbof{}}.}
}
\vspace{-1 ex}
\setlength{\tabcolsep}{4pt}
\begin{tabular}{|l|l|l|cccc|}
\hline
\textbf{Video Scenario} & \textbf{Video Sequence} & \textbf{Description/Challenges} & \textbf{Max \# of people} & \textbf{\# of frames} & \textbf{Resolution} & \textbf{FPS} \\ \hline
\multirow{2}{*}{Common activities} & Lunch meeting 1                                                      & People walking and sitting.                                                                                                               & 11                       & 1,201                & 2,048               & 1          \\ \cline{2-7}
 & Lunch meeting 3                                                      & People walking and sitting.                                                                                                               & 10                       & 900                  & 2,048               & 1          \\ \hline
Crowded scene & Lunch meeting 2                                                      & \begin{tabular}[c]{@{}l@{}}More than 10 people sitting \\ and having lunch.\end{tabular}                                                                                             & 13                       & 3,000                & 2,048               & 10         \\ \hline
Edge Cases & Edge cases                                                           & \begin{tabular}[c]{@{}l@{}}People walking and sitting, \\ extreme body poses, \\ head camouflage, \\ severe body occlusions.\end{tabular}       & 8                        & 4,201                & 2,048               & 10         \\ \hline
Walking activity & High activity                                                        & \begin{tabular}[c]{@{}l@{}}People frequently walking in \\ through one door and leaving \\ through the other door.\end{tabular}              & 9                        & 7,202                & 1,080               & 10         \\ \hline
\multirow{3}{*}[-3em]{Low light} & All-off                                                              & \begin{tabular}[c]{@{}l@{}}People walking and sitting, \\ overhead lights off,\\ camera IR filter removed, \\ no IR illumination\end{tabular}   & 7                        & 3,000                & 1,080               & 10         \\ \cline{2-7}
 & IRfilter                                                             & \begin{tabular}[c]{@{}l@{}}People walking and sitting, \\ overhead lights off,\\ with camera IR filter, \\ no IR illumination\end{tabular}      & 8                        & 3,000                & 1,080               & 10         \\ \cline{2-7}
 & IRill                                                                & \begin{tabular}[c]{@{}l@{}}People walking and sitting, \\ overhead lights off,\\ camera IR filter removed, \\ with IR illumination\end{tabular} & 8                        & 3,000                & 1,080               & 10         \\ \hline
\end{tabular}
\label{table:cepdofdetail}
\end{table*}

\subsection{Dataset}

Although there are several existing datasets for people detection from overhead, fisheye images, either they are not annotated with rotated bounding boxes \cite{mw}, or the number of frames and people are limited \cite{shengye}. Therefore, we collected and labeled a new dataset named Challenging Events for Person Detection from Overhead Fisheye images (\newhabbof{}), and made it publicly available\footnote{\href{http://vip.bu.edu/cepdof}{\tt vip.bu.edu/cepdof}}. We also manually annotated a subset of the \mworiginal{} dataset with rotated bounding-box labels, that we refer to as \mw{}.
We use \mw{}, HABBOF, and \newhabbof{} to evaluate our method and compare it to previous state-of-the-art methods. Table~\ref{table:stats} shows various statistics of these three datasets \zhihaorep{}{, and Table~\ref{table:cepdofdetail} shows the 
{details} of \newhabbof{}}. Clearly, the new \newhabbof{} dataset contains many more frames and human objects, and also includes challenging scenarios such as crowded room, various body poses, and low-light scenarios, which do not exist in the other two datasets. Furthermore, \newhabbof{} is annotated spatio-temporally, i.e., bounding boxes of the same person carry the same ID in consecutive frames, and thus can be also used for additional vision tasks using overhead, fisheye images, such as video-object tracking and human re-identification. 

\begin{table*}[ht] \centering
\caption{Performance comparison of \yolofish{} and previous state-of-the-art methods. P, R, and F denote Precision, Recall, and F-measure, respectively. All metrics are averaged over all the videos in each dataset. Therefore, the F-measure in the table is not equal to the harmonic mean of Precision and Recall results in the table. The inference speed (FPS) is estimated from
a single run on the \textit{Edge cases} video in \newhabbof{} at confidence threshold $\widehat{b}_{\text{conf}} = 0.3$, using Nvidia GTX 1650 GPU.}
\setlength{\tabcolsep}{4pt}
{\small
\begin{tabular}{|l|c|c|ccc|c|ccc|c|ccc|} 
\cline{3-14}
                \multicolumn{2}{c|}{ }                         & \multicolumn{4}{c|}{\mw{}}                                          & \multicolumn{4}{c|}{HABBOF}                                         & \multicolumn{4}{c|}{\newhabbof{}}                                   \\ \hline
                            & FPS          & $\text{AP}_{50}$ & P              & R              & F              & $\text{AP}_{50}$ & P              & R              & F              & $\text{AP}_{50}$ & P              & R              & F              \\ \hline
\hitachi{} \cite{hitachi} (608)   & 6.8          & 78.2             & 0.863          & 0.759          & 0.807          & 87.3             & 0.970          & 0.827          & 0.892          & {61.0}             & {0.884}          & {0.526}          & {0.634}          \\
\shengye\ AA \cite{shengye} {(1,024)} & {0.3}          & 88.4             & 0.939          & 0.819          & 0.874          & {87.7}             & {0.922}          & {0.867}          & {0.892}          & {73.9}             & {0.896}          & {0.638}          & {0.683}          \\
\shengye\ AB \cite{shengye} {(1,024)} & {0.2}          & 95.6             & 0.895          & 0.902          & 0.898          & {93.7}             & {0.881}          & {0.935}          & {0.907}          & {76.9}           & {0.884}          & {0.694}          & {0.743}          \\ \hline
\yolofish{} (608)           & \textbf{7.0} & {96.6}    & {\textbf{0.951}} & {0.931}          & {\textbf{0.941}} & {97.3}             & {\textbf{0.984}} & {0.935}          & {0.958}          & {82.4}             & {\textbf{0.921}} & {0.719}          & {0.793}          \\
\yolofish{} (1,024)          & 3.7          & {\textbf{96.7}}             & {0.919}          & {\textbf{0.951}} & {0.935}          & {\textbf{98.1}}    & {0.975}          & {\textbf{0.963}} & {\textbf{0.969}} & {\textbf{85.8}}    & {0.902}          & {\textbf{0.795}} & {\textbf{0.836}} \\ \hline
\end{tabular}
}
\label{table:oursvsbaseline}
\end{table*}

\subsection{Performance Metrics}

Following the MS COCO challenge \cite{coco}, we adopt Average Precision (AP), i.e., the area under the Precision-Recall curve, as one of our evaluation metrics. However, we only consider AP at IoU = 0.5 {($\text{AP}_{50}$)} since 
even a perfect people-detection algorithm could have a relatively low IoU due to the non-uniqeness of ground truth: for the same person there could be multiple equally good bounding boxes at different angles, but only one of them will be selected by a human annotator to be labeled as the ground truth.
In addition to AP, we also adopt F-measure at a fixed confidence threshold $\widehat{b}_{\text{conf}} = 0.3$ as another performance metric. Note that the F-measure for a given value of $\widehat{b}_{\text{conf}}$ corresponds to a particular point on the Precision-Recall curve.

\newcommand{\vissize}{1.5 in}
\begin{figure*}[t]\centering
    \begin{subfigure}[t]{0.24\textwidth}
        \centering
        \includegraphics[height=\vissize]{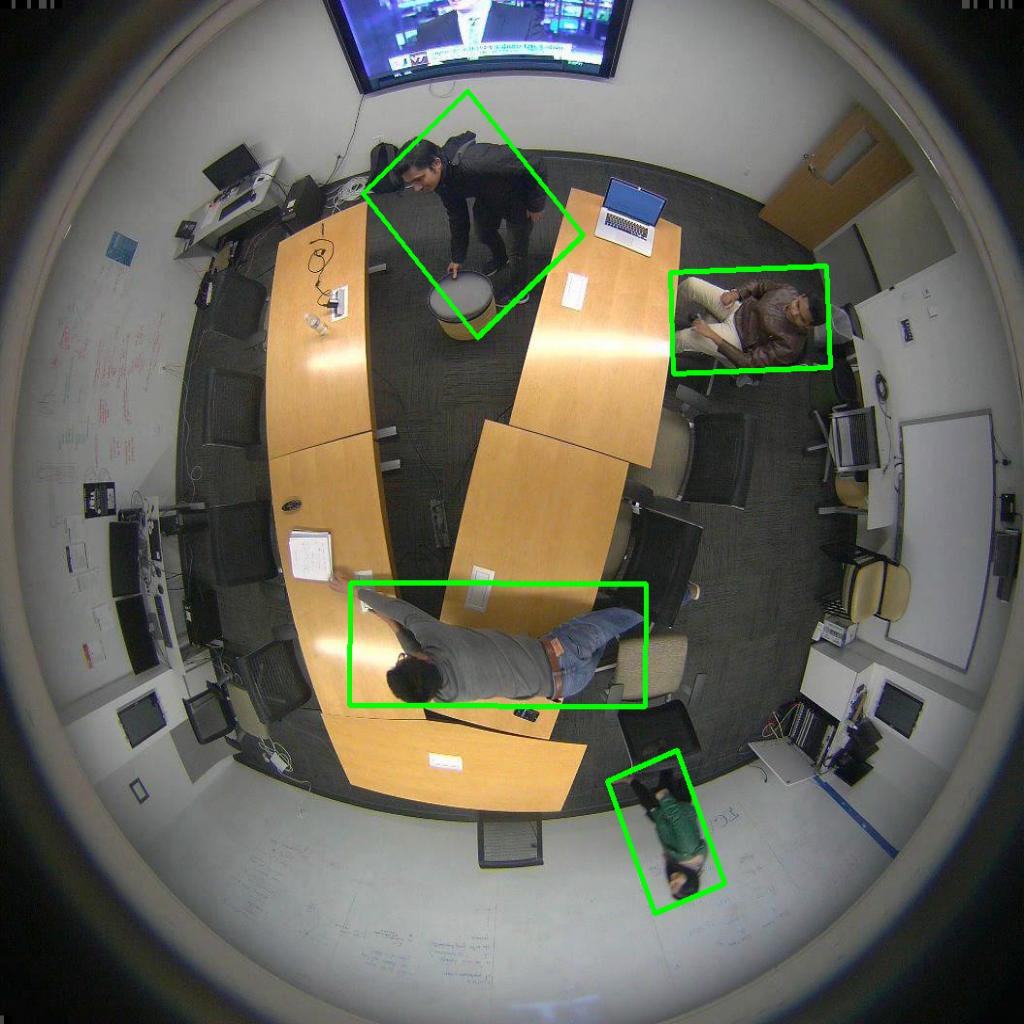}
        \caption{Different poses.}
        \label{fig:imga}
    \end{subfigure}
    \begin{subfigure}[t]{0.24\textwidth}
        \centering
        \includegraphics[height=\vissize]{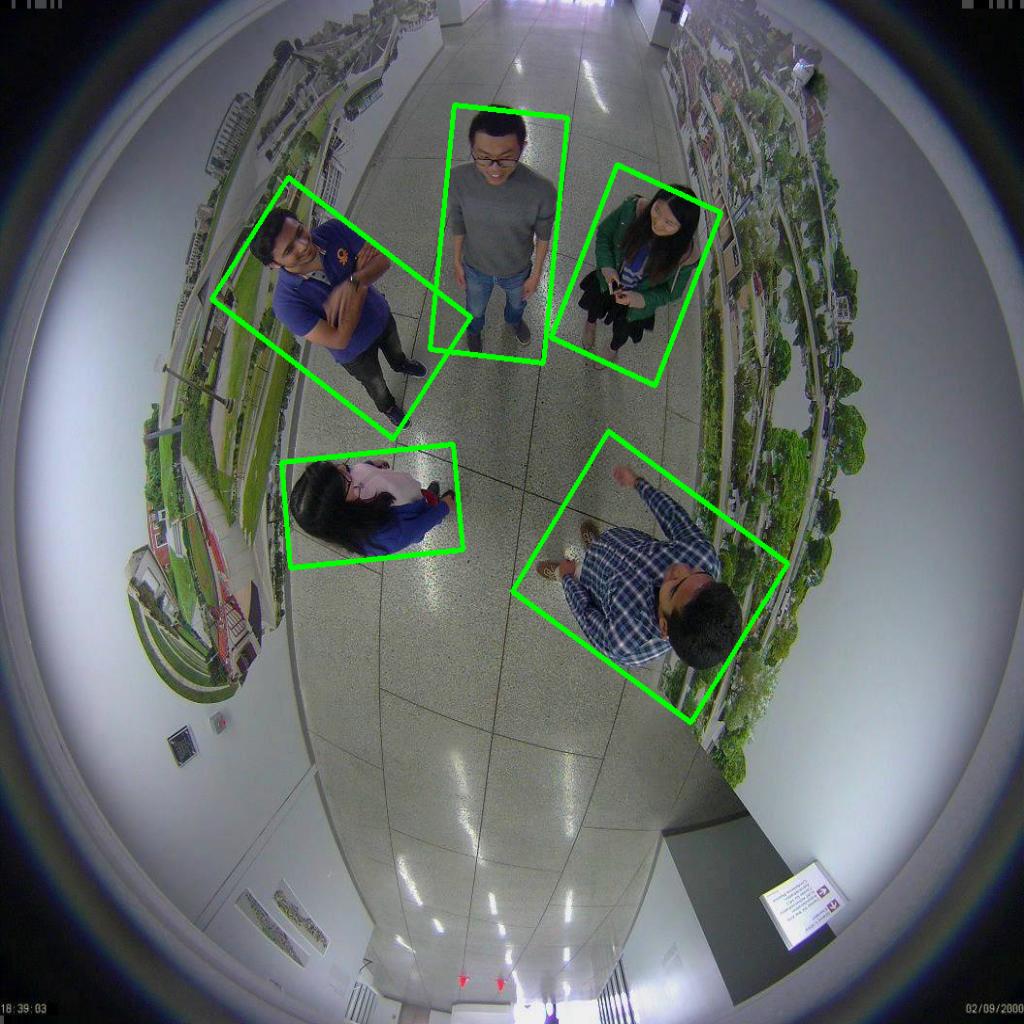}
        \caption{People standing.}
        \label{fig:imgb}
    \end{subfigure}
    \begin{subfigure}[t]{0.24\textwidth}
        \centering
        \includegraphics[height=\vissize]{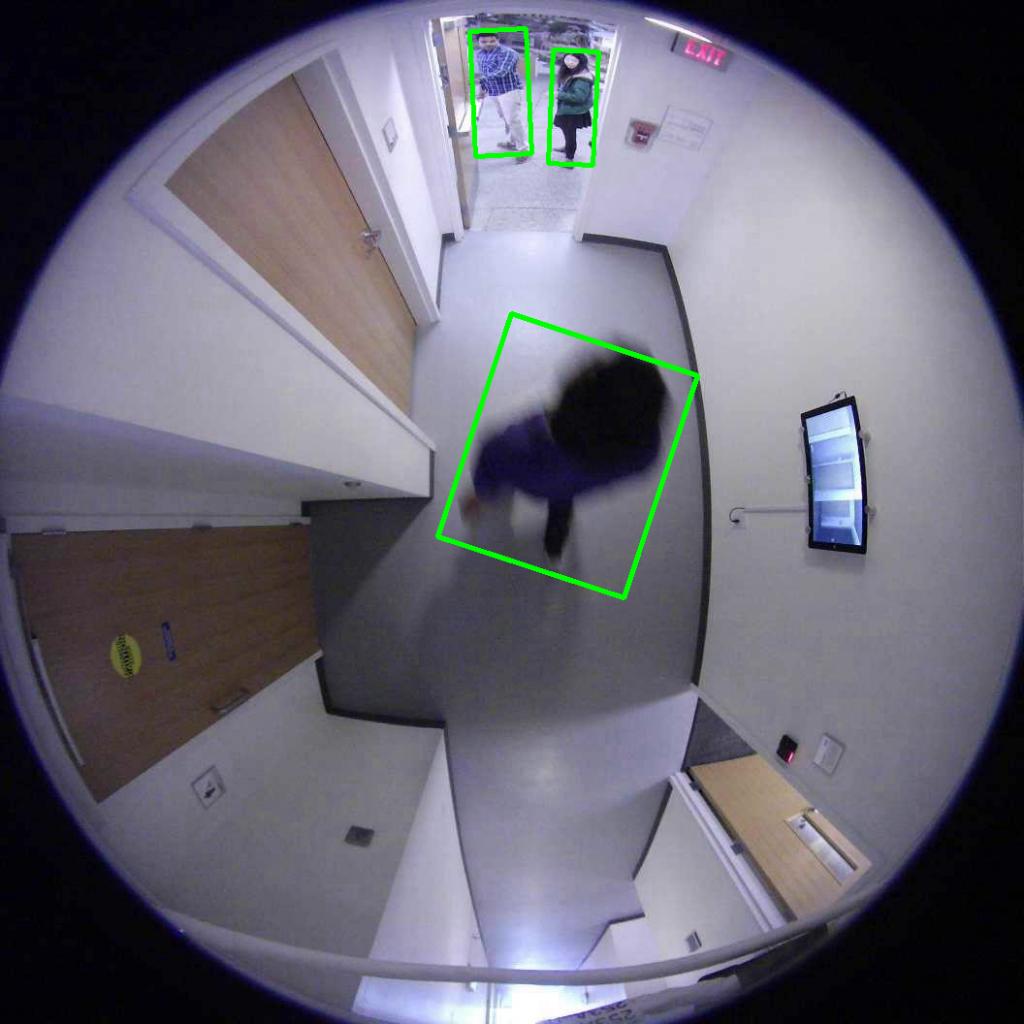}
        \caption{Straight under camera.}
        \label{fig:imgc}
    \end{subfigure}
    \begin{subfigure}[t]{0.24\textwidth}
        \centering
        \includegraphics[height=\vissize]{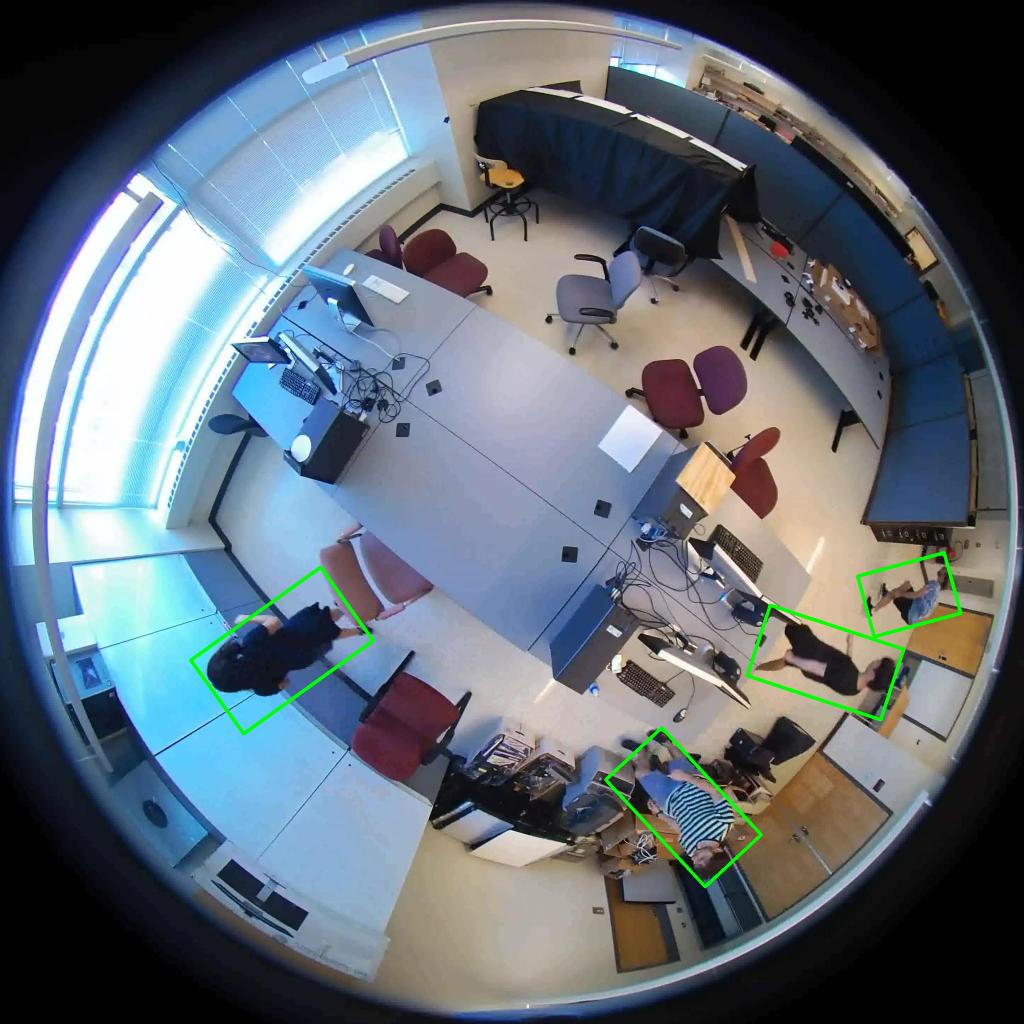}
        \caption{People walking.}
        \label{fig:imgd}
    \end{subfigure}
    \begin{subfigure}[t]{0.24\textwidth}
        \centering
        \includegraphics[height=\vissize]{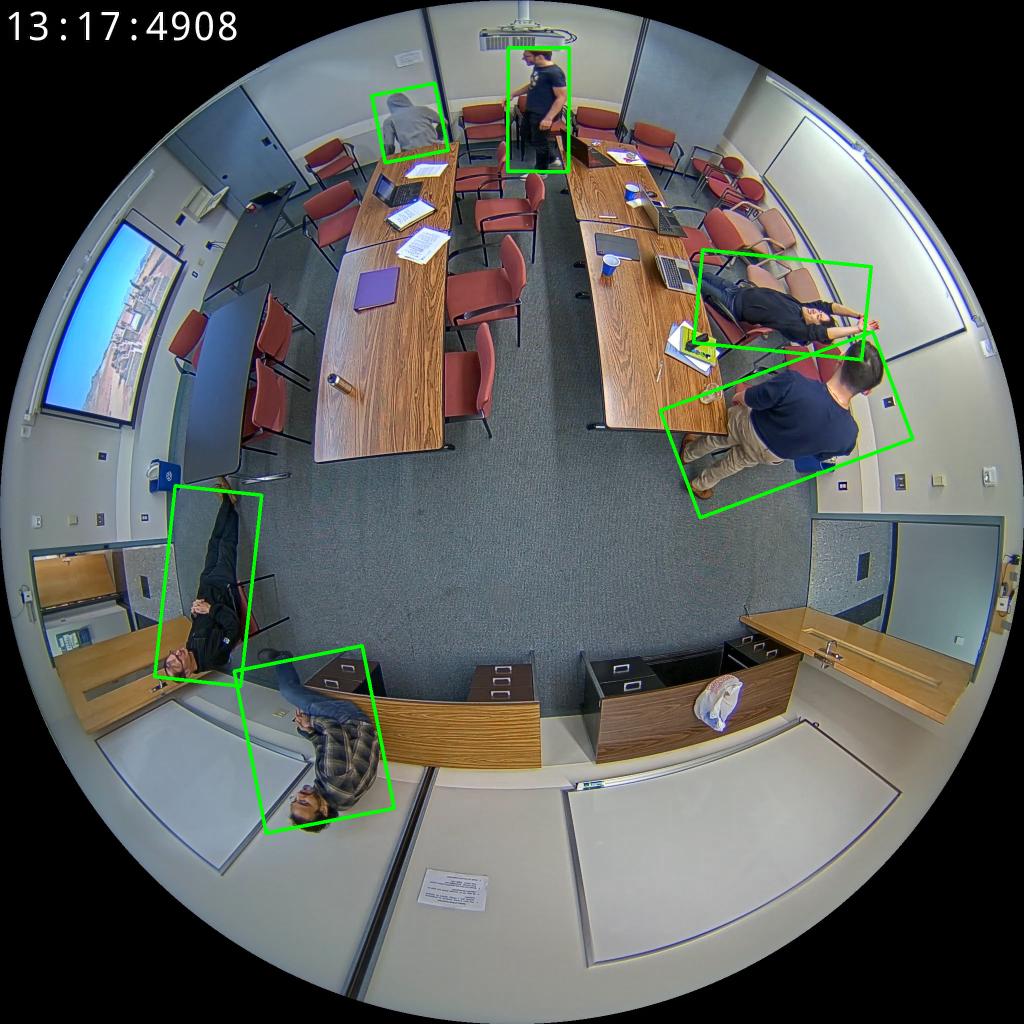}
        \caption{Various angles.}
        \label{fig:imge}
    \end{subfigure}
    \begin{subfigure}[t]{0.24\textwidth}
        \centering
        \includegraphics[height=\vissize]{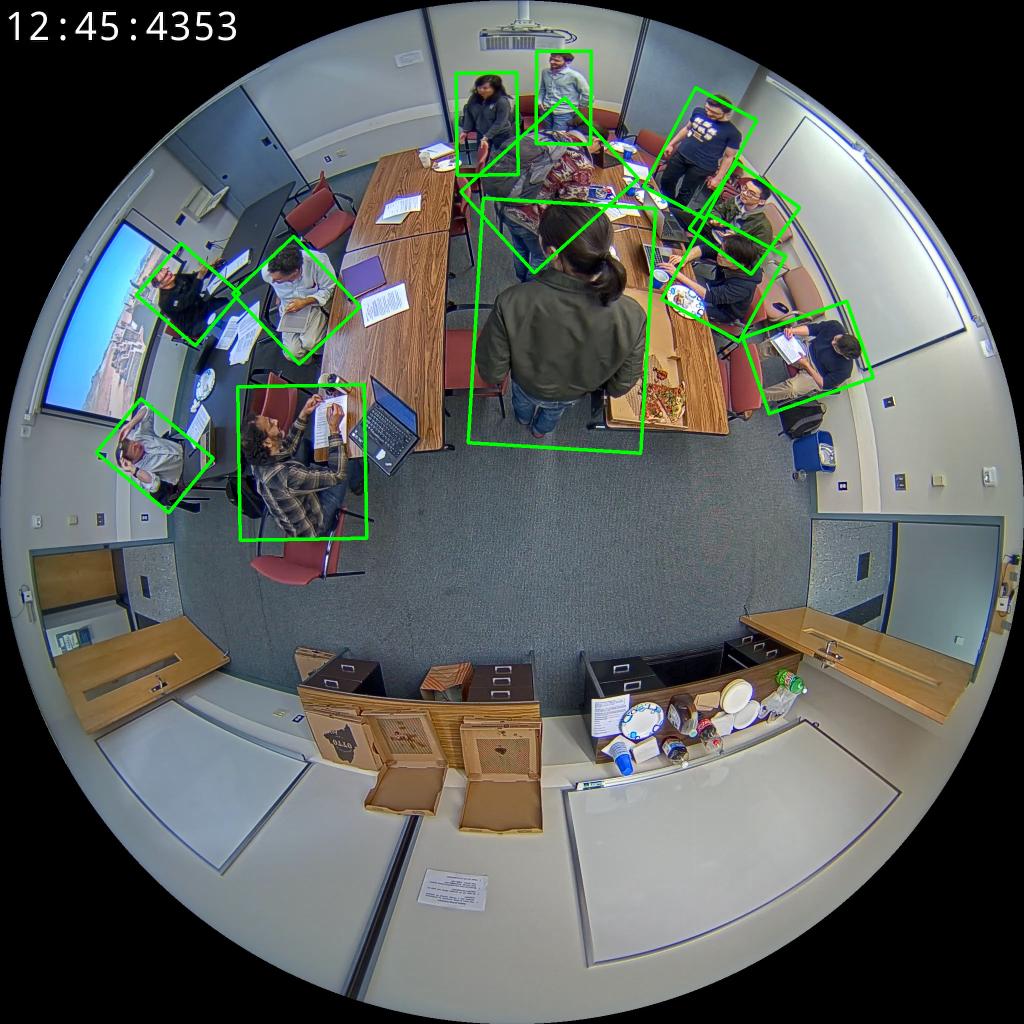}
        \caption{Crowded scene $+$ occlusions.}
        \label{fig:imgf}
    \end{subfigure}
    \begin{subfigure}[t]{0.24\textwidth}
        \centering
        \includegraphics[height=\vissize]{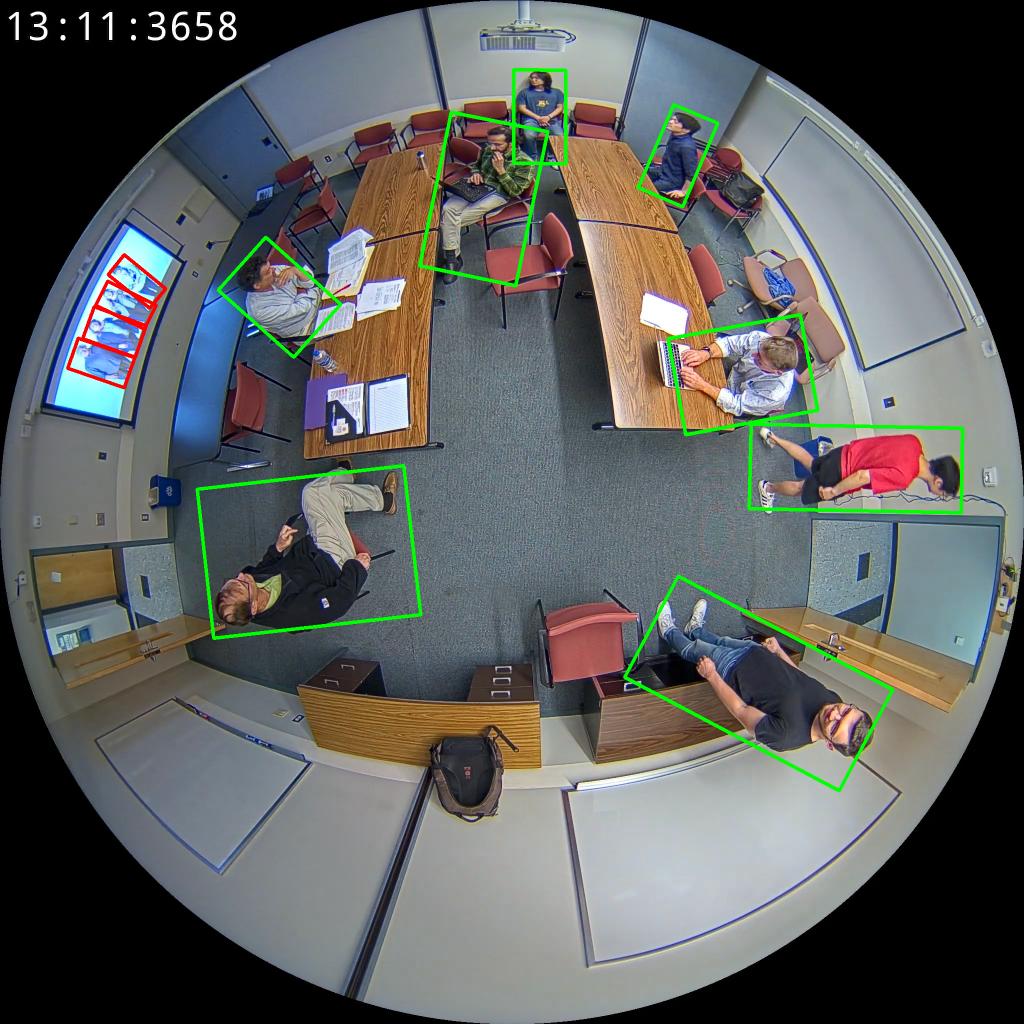}
        \caption{People on the screen.}
        \label{fig:imgg}
    \end{subfigure}
    \begin{subfigure}[t]{0.24\textwidth}
        \centering
        \includegraphics[height=\vissize]{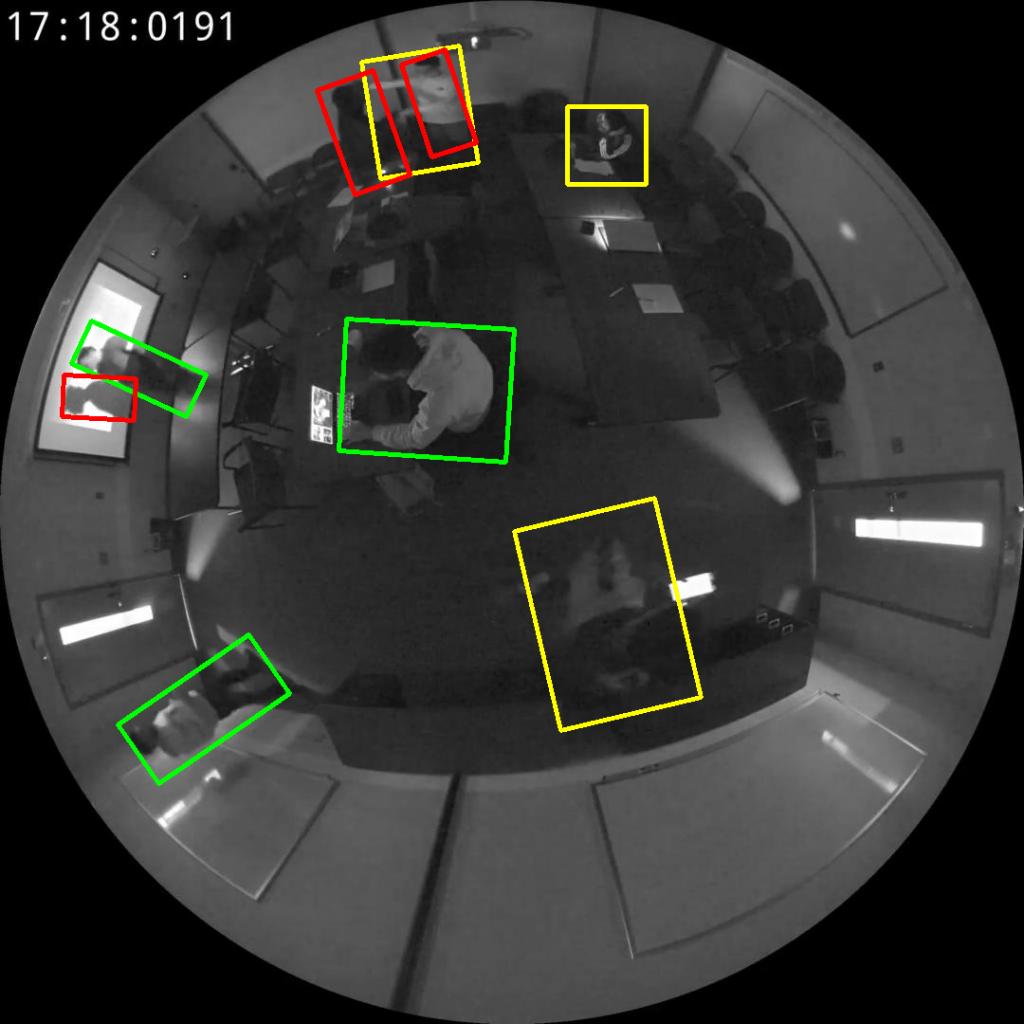}
        \caption{Low-light scenario.}
        \label{fig:imgh}
    \end{subfigure}
    \caption{Qualitative results of \yolofish{} on videos from \mw{} (a--c), HABBOF (d) and \newhabbof{} (e--h). Green boxes are true positives, red boxes are false positives, and yellow boxes are false negatives. Images (a--d) are for relatively easy cases, (e--f) are for challenging cases, and (g--h) are failure examples. As shown in (a--f), \yolofish{} works very well in most scenarios, including various poses, orientations, occupancy levels, and background scenes. However, it produces false positives in (g) on a projection screen {(images of people who should not be counted)} and in (h). It also misses people in low-light conditions, such as in (h).}
    \label{fig:visual}
\end{figure*}

\subsection{Main Results} \label{sec:results}

\textbf{Implementation details:} Unless otherwise specified, we first train our network on the MS COCO 2017 \cite{coco} training images for 100,000 iterations and fine-tune the network on single or multiple datasets from Table~\ref{table:stats} for 6,000 iterations (one iteration contains 128 images). On COCO images, the network weights are updated by Stochastic Gradient Descent (SGD) with the following parameters: step size 0.001, momentum 0.9, and weight decay 
0.0005. For datasets in Table~\ref{table:stats}, we use standard SGD with a step size of 0.0001. 
Rotation, flipping, resizing, and color augmentation are used in both training stages. All results have been computed based on a single run of training and inference.

Table~\ref{table:oursvsbaseline} compares \yolofish{} with other competing algorithms. In order to evaluate AA and AB algorithms from \shengye\ \cite{shengye}, we used the authors' publicly-available implementation.\footnote{\href{http://vip.bu.edu/projects/vsns/cossy/fisheye}{\tt vip.bu.edu/projects/vsns/cossy/fisheye}}
Since the code of \hitachi{} \cite{hitachi} is not publicly available, we implemented their algorithm based on our best understanding.
Since there is no predefined train-test split in these three datasets, we cross-validate \yolofish{} on these datasets, i.e., two datasets are used for training and the remaining one for testing, and this is repeated so that each dataset is used {once} as the test set.
For example, \yolofish{} is trained on \mw{} + HABBOF, and tested on \newhabbof{}, and similarly for other permutations. {We use only one \textit{Low-light} video (with infra-red illumination) during training, as other videos have extremely low contrast, but we use all of them in testing.}
Since neither \shengye{} \cite{shengye} nor \hitachi{} \cite{hitachi} are designed to be trained on rotated bounding boxes, we just trained them on COCO as described in their papers. \hitachi{} {used a} top-view standard-lens image dataset called DPI-T \cite{haque2016dpit} for training in addition to COCO, however currently this dataset is not accessible. In the ablation study (Section~\ref{sec:ablation}), we show the effect of fine-tuning \hitachi{} with overhead, fisheye frames as well.
We use 0.3 as the confidence threshold for all the methods to calculate Precision, Recall, and F-measure.
All methods are tested without test-time augmentation. 

Results in Table~\ref{table:oursvsbaseline} show that \yolofish{} at $608 \times 608$ resolution achieves the best performance and the fastest execution speed among all the methods tested.
Our method is tens of times faster than \shengye's method and slightly faster than the method of \hitachi{}. 
We note that \yolofish{}'s performance is slightly better, {in terms of} AP, than that of \shengye's AB algorithm on the \mw{} dataset in which most human objects {appear in an} upright pose (walking). This is encouraging since people walking or standing appear {radially oriented in} overhead, fisheye images, a scenario for which \hitachi{}'s and \shengye's algorithms have been designed.
However, \yolofish{} outperforms the other algorithms by a large margin on {both} HABBOF, which is relatively easy, and \newhabbof{},
which includes challenging scenarios, such as various body poses and occlusions. We conclude that \yolofish{} works well in both simple and challenging cases while maintaining high computational efficiency. Furthermore, it achieves even better performance when the input image resolution is raised to $1,024 \times 1,024$ but at the cost of a doubled inference time. 
Fig.~\ref{fig:visual} shows sample results of \yolofish{} applied to the three datasets; the detections are nearly perfect in a range of scenarios, such as various body poses, orientations, and diverse background scenes. However, some scenarios, such as people's images on a projection screen (Fig.~\ref{fig:imgg}), low light, and hard shadows, remain challenging.

\subsection{Design Evaluation} \label{sec:design_eval}

We conducted several experiments to analyze the effects of the novel elements we introduced in \yolofish{}. Specifically, we conducted an ablation study and compared different angle loss functions. Due to the limited amount of GPU resources we have, we did not run a full cross-validation for these experiments. Instead, we trained all of these algorithms on COCO and then fine-tuned them on \mw{} using the same optimization parameters as reported in Section~\ref{sec:results}. Then, we tested each algorithm on every video in the HABBOF and \newhabbof{} datasets at $1,024 \times 1,024$ resolution. The resulting AP was averaged over all videos. 

\subsubsection{Ablation Experiments} \label{sec:ablation}

In this section, we present various ablation experiments to analyze how each part of \yolofish{} individually contributes to the overall performance. As the baseline, we use \hitachi\ \cite{hitachi} with NMS and analyze the differences between this baseline and \yolofish{} one-by-one. \hitachi\ use standard YOLO \cite{yolov3} trained on 80-classes of COCO with rotation-invariant training \cite{hitachi} in which the object's angle is uniquely determined by its location.
The first row of Table~\ref{table:ablation} shows the result of this baseline algorithm.
Note that, the baseline algorithm is not trained or fine-tuned on overhead, fisheye frames.

\textbf{Multi-class vs. single-class:}
In \yolofish{}, we remove the category classification part of YOLO since we are dealing with a single object category, namely, person (see Section~\ref{sec:loss}).
As can be seen from the second row of Table~\ref{table:ablation}, this results in a slight performance drop, which is to be expected since training on 80 classes of objects can benefit from multi-task learning.
However, removing the category-classification branch reduces the number of parameters by 0.5M and slightly increases the inference speed (FPS in Table~\ref{table:oursvsbaseline} {and Table~\ref{table:ablation}}).

\textbf{Fine-tuning with overhead, fisheye images:}
To analyze this effect, we fine-tuned the single-class algorithm trained on COCO with images from \mw{}. As shown in the third row of Table~\ref{table:ablation}, this results in a significant performance increase. Recall that the test set used in Table~\ref{table:ablation} does not include any frames from the \mw{} dataset.

\textbf{Rotation-aware people detection:} 
As discussed in Section~\ref{sec:loss}, we introduced a novel loss function to make \yolofish{} \textit{rotation-aware}. Instead of setting the object's angle to be along the FOV radius, we add a parameter, $\widehat{b}_\theta$, to each predicted bounding box and train the network using periodic L1 loss. As shown in the last row of Table~\ref{table:ablation}, the angle prediction further improves the performance of \yolofish{}.

\begin{table}[t]\centering
\caption{Ablation study of \yolofish{}. Fine-tuning is applied using the MW-R dataset.
}
\begin{tabular}{ccc|cc}
\makecell{No.\ of \\ classes} & Angle prediction   & \makecell{Fine-\\tuning} & $\text{AP}_{50}$ & FPS \\ \hline
80                          & Rotation-invariant &                        & {81.4}             & 3.7 \\
1                           & Rotation-invariant &                        & {81.2}             & 3.8 \\
1                           & Rotation-invariant & \checkmark             & {85.9}             & 3.8 \\
1                           & Rotation-aware     & \checkmark             & {\textbf{88.9}}    & 3.7
\end{tabular}
\label{table:ablation}
\end{table}

\begin{table}[t]
\centering
\caption{Comparison of \yolofish{}'s performance for different angle ranges and loss functions.}
\begin{tabular}{cc|c}
Prediction range    & Angle loss  & $\text{AP}_{50}$ \\ \hline
$(-\infty, \infty)$ & L1          &{86.0}              \\
$(-\pi, \pi)$       & L1          & {87.0}             \\
$(-\pi, \pi)$       & Periodic L1 & {\textbf{88.9}}    \\ \hline
$(-\infty, \infty)$ & L2          & {86.1}             \\
$(-\pi, \pi)$       & L2          & {86.1}             \\
$(-\pi, \pi)$       & Periodic L2 & {88.1}            
\end{tabular}
\label{table:angle}
\end{table}

\subsubsection{Comparison of Different Angle Loss Functions} \label{sec:anglelosstable}
To analyze the impact of the loss functions on angle prediction, we ablate the angle value range and angle loss in \yolofish{} while keeping the other parts unchanged. 
{We compare our proposed periodic loss with two baselines: standard unbounded regression loss and bounded regression loss. We {perform the} same experiment for both L1 and L2 loss.} As can be seen in Table~\ref{table:angle}, the periodic L1 loss achieves the best performance, and both the periodic L1 and periodic L2 losses outperform their non-periodic counterparts. 



\subsubsection{Analysis of the Prediction Aspect Ratio} \label{sec:angleloss_hist}
\begin{figure}[t]
\begin{center}
\includegraphics[width=1\linewidth]{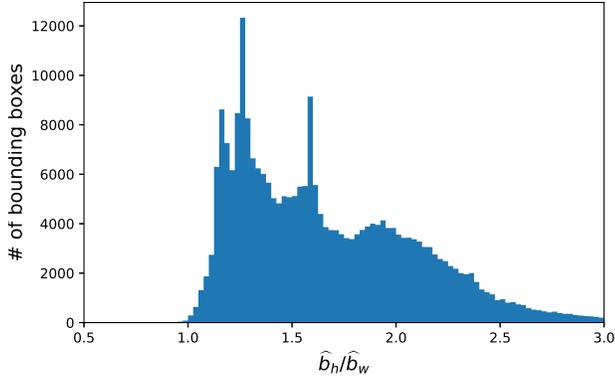}
\end{center}
\vglue -0.5cm
\caption{Histogram of the height-width ratio of the predicted bounding boxes.}
\label{fig:wh}
\end{figure}

As discussed in Section~\ref{sec:angle_range}, we relax the angle range to be inside $[-\pi/2,\pi/2)$ and force $b_w < b_h$ in ground-truth annotations so that every bounding box corresponds to a unique representation.
In the same section, in order to handle the bounding-box symmetry problem we assumed that the network can learn to predict bounding boxes such that $\widehat{b}_w < \widehat{b}_h$.
To demonstrate that this is indeed the case, we analyze the output of our network on both HABBOF and \newhabbof{} datasets. Fig.~\ref{fig:wh} shows the histogram of $\widehat{b}_h / \widehat{b}_w$. We observe that nearly all predicted bounding boxes satisfy $\widehat{b}_w < \widehat{b}_h$ (i.e., $\widehat{b}_h / \widehat{b}_w>1$), which validates our assumption.


\begin{table*}[ht] \centering
\caption{\zhihaorep{}{RAPiD's performance on individual videos in the \newhabbof{} dataset.}}
\setlength{\tabcolsep}{3.5pt}
{\small
\begin{tabular}{|l|c|ccc|c|ccc|c|ccc|c|ccc|}
\hline
             & \multicolumn{4}{c|}{Lunch meeting 1} & \multicolumn{4}{c|}{Lunch meeting 1} & \multicolumn{4}{c|}{Lunch meeting 1}               & \multicolumn{4}{c|}{Edge cases} \\ \hline
             & $\text{AP}_{50}$      & P       & R       & F      & $\text{AP}_{50}$      & P       & R       & F      & \multicolumn{1}{c|}{$\text{AP}_{50}$}    & P     & R     & F     & $\text{AP}_{50}$     & P      & R     & F     \\ \hline
RAPiD (608)  & 96.7   & 0.945   & 0.920   & 0.933  & 95.7   & 0.931   & 0.870   & 0.900  & \multicolumn{1}{c|}{91.3} & 0.905 & 0.794 & 0.846 & 89.2  & 0.967  & 0.791 & 0.870 \\ 
RAPiD (1024) & 97.6   & 0.968   & 0.957   & 0.962  & 97.4   & 0.952   & 0.945   & 0.948  & \multicolumn{1}{c|}{95.3} & 0.896 & 0.885 & 0.891 & 94.0   & 0.949  & 0.901 & 0.925 \\ \hline
             & \multicolumn{4}{c|}{High activity}   & \multicolumn{4}{c|}{All-off}         & \multicolumn{4}{c|}{IRfilter}                      & \multicolumn{4}{c|}{IRill}      \\ \hline
             & $\text{AP}_{50}$      & P       & R       & F      & $\text{AP}_{50}$      & P       & R       & F      & $\text{AP}_{50}$   & P     & R     & F     & $\text{AP}_{50}$     & P      & R     & F     \\ \hline
RAPiD (608)  & 93.2   & 0.966   & 0.884   & 0.923  & 52.8   & 0.853   & 0.428   & 0.570  & 51.4                      & 0.875 & 0.346 & 0.496 & 88.6  & 0.925  & 0.714 & 0.806 \\ 
RAPiD (1024) & 94.2   & 0.964   & 0.913   & 0.938  & 65.6   & 0.798   & 0.532   & 0.638  & 54.5                      & 0.803 & 0.400 & 0.534 & 87.5  & 0.882  & 0.828 & 0.854 \\ \hline
\end{tabular}
}
\label{table:rapiddetail}
\end{table*}

\subsection{
\ozanrep{}{Impact of illumination}
} \label{sec:vid_results}
\zhihaorep{}
{Table~\ref{table:rapiddetail} shows \yolofish{}'s performance for each video in the \newhabbof{} dataset. Clearly, when running at $1,024\times 1,024$-pixel resolution RAPiD performs extremely well on normal-light videos (\textit{Lunch meeting 1/2/3}, \textit{Edge cases}, and \textit{High activity}) with $\text{AP}_{50} \geq 94.0$. However, both $\text{AP}_{50}$ and F-measure drop significantly for \textit{All-off} and \textit{IRfilter} videos. We observe that RAPiD has a relatively high Precision but very low Recall on these two videos, \ie, it misses many people. By comparing RAPiD's output with ground-truth annotations, we find that the people RAPiD misses are usually indistinguishable from the background
{(see Fig.~\ref{fig:imgh} for an example)}. Detecting such barely-visible people from a single video frame is a very challenging task even for humans. Notably, when IR illumination is turned on ({\it IRill} video), RAPiD's performance vastly improves to levels only slightly sub-par compared to that for normal-light videos.}

\section{Conclusions}
In this paper, we proposed \yolofish{}, a novel people detection algorithm for overhead, fisheye images. Our algorithm extends object-detection algorithms which use axis-aligned bounding boxes, such as YOLO, to the case of person detection using human-aligned bounding boxes. We show that our proposed periodic loss function outperforms traditional regression loss functions in angle prediction. With rotation-aware bounding box prediction, \yolofish{} outperforms previous state-of-the-art methods by a large margin without introducing additional computational complexity. 
{Unsurprisingly, RAPiD's performance drops significantly for videos captured in extremely low-light scenarios, where people are barely distinguishable from the background. Further research is needed to address such scenarios.}
We also introduced a new dataset, that consists of 25K frames and 173K people annotations. We believe both our method and dataset will be beneficial for various real-world applications and research using overhead, fisheye images and videos.

{\small
\bibliographystyle{ieee_fullname}
\bibliography{strings, main}

\begin{thebibliography}{10}\itemsep=-1pt

\bibitem{brunetti2018survey_deep}
Antonio Brunetti, Domenico Buongiorno, Gianpaolo~Francesco Trotta, and
  Vitoantonio Bevilacqua.
\newblock Computer vision and deep learning techniques for pedestrian detection
  and tracking: A survey.
\newblock {\em Neurocomputing}, 300:17--33, 2018.

\bibitem{Chiang2014HumanDI}
An-Ti Chiang and Yao Wang.
\newblock Human detection in fish-eye images using hog-based detectors over
  rotated windows.
\newblock {\em Proc.\ IEEE Intern.\ Conf.\ on Multimedia and Expo Workshops},
  pages 1--6, 2014.

\bibitem{dalal2005hog}
Navneet Dalal and Bill Triggs.
\newblock Histograms of oriented gradients for human detection.
\newblock In {\em Proc.\ IEEE Conf.\ Computer Vision Pattern Recognition},
  volume~1, pages 886--893. IEEE, 2005.

\bibitem{roitransformer}
Jian Ding, Nan Xue, Yang Long, Gui-Song Xia, and Qikai Lu.
\newblock Learning roi transformer for oriented object detection in aerial
  images.
\newblock In {\em Proc.\ IEEE Conf.\ Computer Vision Pattern Recognition},
  pages 2849--2858, 2019.

\bibitem{dollar2014acf}
Piotr Doll{\'a}r, Ron Appel, Serge Belongie, and Pietro Perona.
\newblock Fast feature pyramids for object detection.
\newblock {\em IEEE Trans.\ Pattern Anal.\ Machine Intell.}, 36(8):1532--1545,
  2014.

\bibitem{enzweiler2008survey}
Markus Enzweiler and Dariu~M Gavrila.
\newblock Monocular pedestrian detection: Survey and experiments.
\newblock {\em IEEE Trans.\ Pattern Anal.\ Machine Intell.}, 31(12):2179--2195,
  2008.

\bibitem{dssd}
Cheng-Yang {Fu}, Wei {Liu}, Ananth {Ranga}, Ambrish {Tyagi}, and Alexander~C.
  {Berg}.
\newblock {DSSD : Deconvolutional Single Shot Detector}.
\newblock {\em arXiv e-prints}, page arXiv:1701.06659, Jan 2017.

\bibitem{fastrcnn}
Ross Girshick.
\newblock Fast r-cnn.
\newblock In {\em Proc.\ IEEE Int.\ Conf.\ Computer Vision}, December 2015.

\bibitem{haque2016dpit}
Albert Haque, Alexandre Alahi, and Li Fei-Fei.
\newblock Recurrent attention models for depth-based person identification.
\newblock In {\em Proc.\ IEEE Conf.\ Computer Vision Pattern Recognition},
  pages 1229--1238, 2016.

\bibitem{maskrcnn}
Kaiming He, Georgia Gkioxari, Piotr Dollar, and Ross Girshick.
\newblock Mask r-cnn.
\newblock In {\em Proc.\ IEEE Int.\ Conf.\ Computer Vision}, Oct 2017.

\bibitem{Krams2017PeopleDI}
O. {Krams} and N. {Kiryati}.
\newblock People detection in top-view fisheye imaging.
\newblock In {\em Proc.\ IEEE Int.\ Conf.\ Advanced Video and Signal-Based
  Surveillance}, pages 1--6, Aug 2017.

\bibitem{shengye}
S. {Li}, M.~O. {Tezcan}, P. {Ishwar}, and J. {Konrad}.
\newblock Supervised people counting using an overhead fisheye camera.
\newblock In {\em Proc.\ IEEE Int.\ Conf.\ Advanced Video and Signal-Based
  Surveillance}, pages 1--8, Sep. 2019.

\bibitem{fpn}
Tsung-Yi Lin, Piotr Doll{\'a}r, Ross Girshick, Kaiming He, Bharath Hariharan,
  and Serge Belongie.
\newblock Feature pyramid networks for object detection.
\newblock In {\em Proc.\ IEEE Conf.\ Computer Vision Pattern Recognition},
  pages 2117--2125, 2017.

\bibitem{coco}
Tsung-Yi {Lin}, Michael {Maire}, Serge {Belongie}, Lubomir {Bourdev}, Ross
  {Girshick}, James {Hays}, Pietro {Perona}, Deva {Ramanan}, C.~Lawrence
  {Zitnick}, and Piotr {Doll{\'a}r}.
\newblock {Microsoft COCO: Common Objects in Context}.
\newblock {\em arXiv e-prints}, page arXiv:1405.0312, May 2014.

\bibitem{ssd}
Wei {Liu}, Dragomir {Anguelov}, Dumitru {Erhan}, Christian {Szegedy}, Scott
  {Reed}, Cheng-Yang {Fu}, and Alexander~C. {Berg}.
\newblock {SSD: Single Shot MultiBox Detector}.
\newblock {\em arXiv e-prints}, page arXiv:1512.02325, Dec 2015.

\bibitem{ma2018rrpn}
Jianqi Ma, Weiyuan Shao, Hao Ye, Li Wang, Hong Wang, Yingbin Zheng, and
  Xiangyang Xue.
\newblock Arbitrary-oriented scene text detection via rotation proposals.
\newblock {\em IEEE Trans.\ Multimedia}, 20(11):3111--3122, 2018.

\bibitem{mw}
Nuo Ma.
\newblock Mirror worlds challenge.

\bibitem{nguyen2016survey}
Duc~Thanh Nguyen, Wanqing Li, and Philip~O Ogunbona.
\newblock Human detection from images and videos: A survey.
\newblock {\em Pattern Recognition}, 51:148--175, 2016.

\bibitem{nosaka2018orientation}
Ryusuke Nosaka, Hidenori Ujiie, and Takaharu Kurokawa.
\newblock Orientation-aware regression for oriented bounding box estimation.
\newblock In {\em Proc.\ IEEE Int.\ Conf.\ Advanced Video and Signal-Based
  Surveillance}, pages 1--6. IEEE, 2018.

\bibitem{qian2019rsdet}
Wen Qian, Xue Yang, Silong Peng, Yue Guo, and Chijun Yan.
\newblock Learning modulated loss for rotated object detection.
\newblock {\em arXiv preprint arXiv:1911.08299}, 2019.

\bibitem{yolo}
Joseph Redmon, Santosh Divvala, Ross Girshick, and Ali Farhadi.
\newblock You only look once: Unified, real-time object detection.
\newblock In {\em Proc.\ IEEE Conf.\ Computer Vision Pattern Recognition}, June
  2016.

\bibitem{yolov2}
Joseph Redmon and Ali Farhadi.
\newblock Yolo9000: Better, faster, stronger.
\newblock In {\em Proc.\ IEEE Conf.\ Computer Vision Pattern Recognition}, July
  2017.

\bibitem{yolov3}
Joseph Redmon and Ali Farhadi.
\newblock Yolov3: An incremental improvement.
\newblock {\em arXiv}, 2018.

\bibitem{fasterrcnn}
Shaoqing Ren, Kaiming He, Ross Girshick, and Jian Sun.
\newblock Faster r-cnn: Towards real-time object detection with region proposal
  networks.
\newblock In C. Cortes, N.~D. Lawrence, D.~D. Lee, M. Sugiyama, and R. Garnett,
  editors, {\em Proc.\ Conf.\ Neural Inf.\ Proc.\ Systems}, pages 91--99.
  Curran Associates, Inc., 2015.

\bibitem{saito2011people}
Mamoru Saito, Katsuhisa Kitaguchi, Gun Kimura, and Masafumi Hashimoto.
\newblock People detection and tracking from fish-eye image based on
  probabilistic appearance model.
\newblock In {\em SICE Annual Conference 2011}, pages 435--440. IEEE, 2011.

\bibitem{seidel2018improved}
Roman Seidel, Andr{\'e} Apitzsch, and Gangolf Hirtz.
\newblock Improved person detection on omnidirectional images with non-maxima
  suppression.
\newblock {\em arXiv preprint arXiv:1805.08503}, 2018.

\bibitem{hitachi}
Masato Tamura, Shota Horiguchi, and Tomokazu Murakami.
\newblock Omnidirectional pedestrian detection by rotation invariant training.
\newblock In {\em Proc.\ IEEE Winter Conf.\ on Appl.\ of Computer Vision},
  pages 1989--1998. IEEE, 2019.

\bibitem{efficientdet}
Mingxing {Tan}, Ruoming {Pang}, and Quoc~V. {Le}.
\newblock {EfficientDet: Scalable and Efficient Object Detection}.
\newblock {\em arXiv e-prints}, page arXiv:1911.09070, Nov 2019.

\bibitem{fcos}
Zhi Tian, Chunhua Shen, Hao Chen, and Tong He.
\newblock Fcos: Fully convolutional one-stage object detection.
\newblock In {\em Proc.\ IEEE Int.\ Conf.\ Computer Vision}, October 2019.

\bibitem{Wang2017TemplateBP}
T. {Wang}, C. {Chang}, and Y. {Wu}.
\newblock Template-based people detection using a single downward-viewing
  fisheye camera.
\newblock In {\em Intern.\ Symp.\ on Intell.\ Signal Process.\ and Comm.\
  Systems}, pages 719--723, Nov 2017.

\bibitem{r3det}
Xue Yang, Qingqing Liu, Junchi Yan, and Ang Li.
\newblock R3det: Refined single-stage detector with feature refinement for
  rotating object.
\newblock {\em arXiv preprint arXiv:1908.05612}, 2019.

\bibitem{atss}
Shifeng {Zhang}, Cheng {Chi}, Yongqiang {Yao}, Zhen {Lei}, and Stan~Z. {Li}.
\newblock {Bridging the Gap Between Anchor-based and Anchor-free Detection via
  Adaptive Training Sample Selection}.
\newblock {\em arXiv e-prints}, page arXiv:1912.02424, Dec 2019.

\bibitem{m2det}
Qijie {Zhao}, Tao {Sheng}, Yongtao {Wang}, Zhi {Tang}, Ying {Chen}, Ling {Cai},
  and Haibin {Ling}.
\newblock {M2Det: A Single-Shot Object Detector based on Multi-Level Feature
  Pyramid Network}.
\newblock {\em arXiv e-prints}, page arXiv:1811.04533, Nov 2018.

\bibitem{zhou2017orientationawarecnn}
Yanzhao Zhou, Qixiang Ye, Qiang Qiu, and Jianbin Jiao.
\newblock Oriented response networks.
\newblock In {\em Proc.\ IEEE Conf.\ Computer Vision Pattern Recognition},
  pages 519--528, 2017.

\end{thebibliography}
}

\clearpage

\onecolumn

\begin{appendices}



\end{appendices}

\end{document}